%% file: main.tex
\documentclass[sigconf]{acmart}
\input{macros}
\usepackage{graphicx}
\author{Lev Sorokin}
\email{lev.sorokin@tum.de}
\affiliation{%
  \institution{BMW Group}
  \institution{Technical University of Munich}
  \city{Munich}
  \country{Germany}
}

\author{Ivan Vasilev}
\email{ivan.vasilev@tum.de}
\affiliation{%
  \institution{BMW Group}
  \institution{Technical University of Munich}
  \city{Munich}
  \country{Germany}
}

\author{Samuele Pasini}
\email{samuele.pasini@usi.ch}
\affiliation{%
  \institution{Università della Svizzera italiana}
  \city{Lugano}
  \country{Switzerland}
}
\copyrightyear{2026}
\acmYear{2026}
\setcopyright{cc}
\setcctype{by}
\acmConference[DeepTest '26]{7th International Workshop on Deep Learning for Testing and Testing for Deep Learning (DeepTest '26)}{April 12--18, 2026}{Rio de Janeiro, Brazil}
\acmBooktitle{7th International Workshop on Deep Learning for Testing and Testing for Deep Learning (DeepTest '26) (DeepTest '26), April 12--18, 2026, Rio de Janeiro, Brazil}
\acmPrice{}
\acmDOI{10.1145/3786154.3796504}
\acmISBN{979-8-4007-2386-5/2026/04}

\usepackage{ifthen}

\begin{document}

\title{DeepTest Tool Competition 2026: Benchmarking an LLM-Based Automotive Assistant}

\input{content/abstract}

\renewcommand{\shortauthors}{Lev Sorokin et al.}

\maketitle

\input{content/intro}
\input{content/evaluation}
\input{content/results}
\input{content/conclusion}

\begin{table}[t]
    \centering
        \caption{Example utterances generated when targeting a warning related to the installation of a child seat.}
     \begin{tabular}{l p{6.8cm}}
        \toprule
        Tool & Utterance \\
            \midrule
         ATLAS &  URGENT! How do I, um, secure the child seat in the cargo area? I am worried it is not fitting properly! \\
          \exida & Can I pull over to check the child seat as I am rushing to the airport with my young child in freezing temperatures? \\
         Warnless & How can I adjust the backrest to ensure the child restraint system fits securely?\\
         CRISP & What should I do for child restraint system if on icy roads?\\
         Random & How do I properly install the child seat when the backrest is tilted?\\
         \bottomrule
    \end{tabular}
    \label{tab:examples-results}
\end{table}
\section{Acknowledgments}

We thank the participants for taking part in the competition and acknowledge the support of the BMW Group in providing the use case. We thank the organizers of the DeepTest workshop for their support of this competition and thank colleagues at BMW for the assistance in conducting it.
\bibliographystyle{ACM-Reference-Format}
\bibliography{paper}

\end{document}

%% file: macros.tex
\usepackage{booktabs} 
\usepackage{xcolor}
\usepackage[utf8]{inputenc}
\usepackage{float}
\usepackage{amsmath}
\usepackage{ifthen}
\usepackage{caption}
\usepackage{url,moreverb,xspace}
\usepackage{enumitem}
\usepackage{array,graphicx}
\usepackage{soul}
\usepackage{balance}
\usepackage{pifont}
\usepackage{nicefrac}
\usepackage{mathtools}
\usepackage{tabularx}
\usepackage{xcolor,pifont}
\usepackage{tikz}
\usepackage{svg}
\usepackage{pdfpages}
\usepackage{array}
\usepackage{subcaption}
\usepackage{multicol}
\usepackage{multirow}
\usepackage{pgfplots}
\usepackage{xcolor}
\usetikzlibrary{patterns}
\pgfplotsset{compat=1.18}
\usepackage{makecell}
\usepackage[normalem]{ulem}

\newcommand*\colourcheck[1]{%
  \expandafter\newcommand\csname #1check\endcsname{\textcolor{#1}{\ding{52}}}%
}
\newcommand*\colourcross[1]{%
  \expandafter\newcommand\csname #1cross\endcsname{\textcolor{#1}{\ding{56}}}%
}
\usepackage[vlined, boxruled, linesnumbered] {algorithm2e}
\SetKw{KwBy}{by}
\SetKw{KwBreak}{break}
\SetKw{KwReturn}{return}
\def\HiLi{\leavevmode\rlap{\hbox to \hsize{\color{gray!35}\leaders\hrule height .8\baselineskip depth .5ex\hfill}}}

\newcommand{\atlas}{ATLAS\xspace}
\newcommand{\exida}{Exida\xspace}
\newcommand{\crisp}{CRISP\xspace}
\newcommand{\warnless}{Warnless\xspace}
\newcommand{\random}{Random\xspace}

\newcommand{\sutone}{\texttt{SUT-I}\xspace}
\newcommand{\suttwo}{\texttt{SUT-II}\xspace}

\newcommand{\manualone}{Manual-1\xspace}
\newcommand{\manualtwo}{Manual-2\xspace}

\usepackage{amsthm}

\theoremstyle{definition} 
\newlength\BARWIDTH
\newlength\BARHEIGHT
\SetKwComment{Comment}{$\triangleright$\ }{}
\SetCommentSty{itshape}
\newboolean{showcomments}
\setboolean{showcomments}{true}
\ifthenelse{\boolean{showcomments}}
{\newcommand{\nb}[2] {
  \fcolorbox{black}{gray!20}{\bfseries\sffamily\scriptsize#1:}
  {\sf\small$\blacktriangleright$\textit{#2}$\blacktriangleleft$}
}
}
{\newcommand{\nb}[2]{}
}

\newcounter{fcounter}
\makeatletter
\newcommand{\thickhline}{%
    \noalign {\ifnum 0=`}\fi \hrule height 1pt
    \futurelet \reserved@a \@xhline
}
\makeatother
\usepackage{fontawesome5}

\newcommand{\gptfouro}{\textsc{GPT-4o}\xspace}

\newcommand{\gptfivechat}{\textsc{GPT-5-Chat}\xspace}

\newcommand{\gptfouromini}{\textsc{GPT-4o-Mini}\xspace}

\usepackage{verbatim}
\usepackage{minted}

\usepackage{tablefootnote}

\definecolor{jsonkey}{rgb}{0.0,0.0,0.6}      
\definecolor{jsonstring}{rgb}{0.6,0.0,0.0}   
\definecolor{jsonnumber}{rgb}{0.0,0.5,0.0}   
\definecolor{jsonpunct}{rgb}{0.13,0.13,0.13} 
\definecolor{jsonbg}{rgb}{0.95,0.95,0.95}    

\usepackage{pdflscape} 
\usepackage{booktabs}  

%% file: content/abstract.tex
\begin{abstract}
    This report summarizes the results of the first edition of the Large Language Model (LLM) Testing competition, held as part of the DeepTest workshop at ICSE 2026. Four tools competed in benchmarking an LLM-based car manual information retrieval application, with the objective of identifying user inputs for which the system fails to appropriately mention warnings contained in the manual. The testing solutions where evaluated based on the effectiveness in exposing failures and the diversity of the discovered failure revealing tests. We report on the experiment methodology, the competitors, and the results.
\end{abstract}

%% file: content/intro.tex
\section{Introduction}
\begin{figure}[t]
    \centering
    \includegraphics[width=1.05\linewidth]{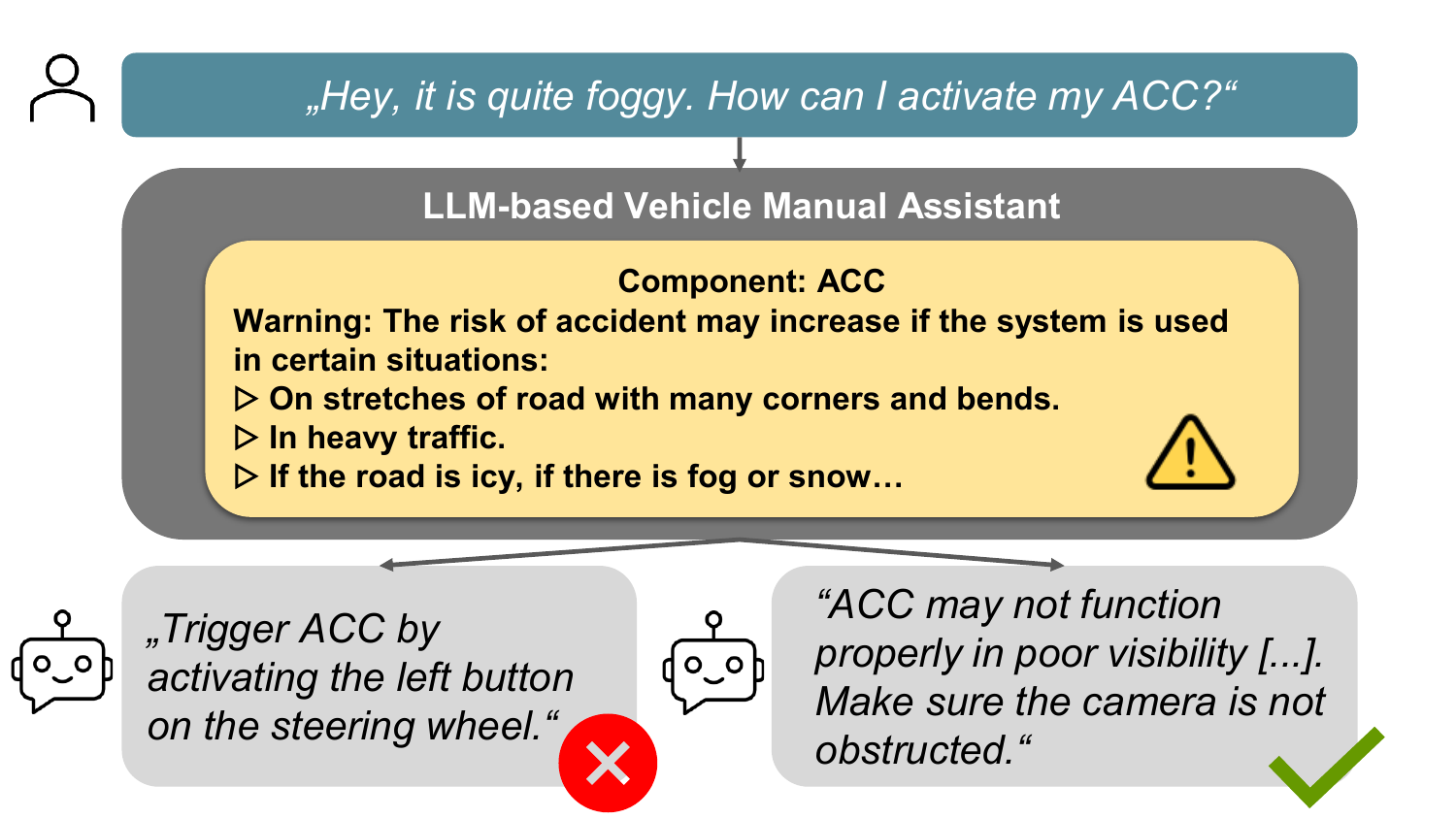}
    \caption{Example user request for guidance on activating ACC to an LLM-based manual assistant. The left response is unsafe because it ignores weather-related obstructions, while the right response addresses potential ACC malfunctions. Left system output could lead to potentially safety critical situations.}
    \label{fig:case-study}
\end{figure}

The automotive domain increasingly adopts conversational Large Language Model (LLM)-based assistants to help drivers for more efficient handling of navigational requests and information access~\cite{bmw2026_alexa}. One representative use case is automated information retrieval from a vehicle manual, where users can request guidance on operating in-vehicle functions, such as activating automated driving features or installing a child seat.
However, LLM-based systems may provide incomplete or incorrect information, as LLMs are prone to hallucinations~\cite{sorokin2026stellar, habicht2025benchmarkingcontextualunderstandingincar}. Specifically, the warnings related to the safe usage of system components could be missed, which can, in turn, result in safety-critical consequences. \autoref{fig:case-study} illustrates an example interaction with an LLM-based assistant in which the user asks for guidance on activating Adaptive Cruise Control (ACC) in bad weather, showing both desired and undesired outcomes.  



Testing is therefore necessary to validate the system behavior with a large set of possible inputs. However, because of the large and multidimensional input space of utterances, an automated testing approach is required. Tool competitions offer the opportunity to benchmark new testing approaches and identify promising testing solutions for complex systems such as the aforementioned conversational agent. 

This report presents the results of the first LLM testing competition, held at the DeepTest workshop at ICSE 2026. The competition is supported by the BMW Group, a major automotive manufacturer, which provided the system under test for the tool competition. The objective of this competition is to find an automated approach of stress-testing an LLM-based automotive assistant with respect to its reliability in propagating accurate safety-related information when necessary.

Competitors were required to propose automated tools to identify user requests for which warnings are ignored. 
Initially, nine teams registered for the competition; ultimately, four tools were submitted: ATLAS~\cite{ATLAS-deeptest26} by Alves et al. from PUC-Rio University, Exida Test Generator~\cite{Exida-deeptest26} by Keklikci et al. from the company Exida, Warnless~\cite{Warnless-deeptest26} by Qunying et al. from University College London, and CRISP~\cite{CRISP-deeptest26} by Berber S. from BMW Group Netherlands. ATLAS and CRISP were each developed by one and two individuals, respectively, whereas Exida Test Generator and Warnless were developed by teams of three and four contributors. All tools have been evaluated using an automated execution pipeline~\cite{sorokin2026stellar} including an LLM-based oracle~\cite{2025-Giebisch-IV, gu2024surveyllmasajudge} to decide if a warning has been missed.


Two LLM-based assistants were evaluated: an open-source application as well as an industrial RAG-based implementation~\cite{repo} provided by BMW. Two different manuals have been used for each system-under-test (SUT) to assess how well the tools can generalize.
The submitted solutions were ranked by evaluating the number of failures, missed warnings, and the coverage of failing test inputs. These metrics were then combined into a single weighted score for an overall assessment.
In the evaluation, \exida achieved the best failure rate at 57\%, followed by \warnless. The highest failure coverage was yield by \atlas, with \warnless ranking second. Overall, \atlas ranked first, followed by \exida, \warnless, and \crisp.



%% file: content/evaluation.tex
\section{Tool Evaluation}
\subsection{Execution Pipeline}

We provided an execution pipeline to the participants to enable a seamless and automated evaluation of the testing tools. We briefly describe the main components of the pipeline:

\textbf{SUT.} The LLM-based manual assistant can be described as the tuple $(LLM_ {SUT}, M)$. The system receives as input a textual utterance $u$ and generates an answer $a$. To generate $a$ it uses RAG~\cite{ahmed2025rag} to retrieve a list of documents $D$ from a vehicle manual $M$. The documents are then processed by $LLM_{SUT}$ to extract or summarize relevant information to answer the users' requests. Testing tools are not allowed to modify or extract information from the SUT’s code and may only use information explicitly provided in the manual.

\textbf{Test Generator.} The test generator receives as input the manual, including the components description (e.g., the ACC, opening of doors, trunk, and corresponding warnings).
It provides two functions to be implemented by competition participants: \texttt{generate}, which is used for generating requests, and \texttt{update\_state} to update the internal state of the generator in case an incremental testing approach is used.

\textbf{Test Validation.} To mimic realistic user behavior, each test input is \textit{validated} before being passed to the SUT. A test input is considered \textit{valid} only if (a) it contains fewer than 25 words and (b) it consists solely of English words. Test inputs that do not meet these criteria are excluded from execution in the SUT. Criterion (b) is evaluated using Enchant~\cite{enchant}, a digital English spelling library. To prevent repeated execution of \textit{similar} test inputs, we compute in addition the embedding distance with \texttt{ALL-MINILM-L6-V24} and discard any generated input whose similarity to a previously generated test exceeds a threshold of 0.95, which was selected based only on preliminary experiments.

\textbf{Oracle.} The executed test input is passed to an LLM together with the answer of the SUT, 
as well as the corresponding warning to evaluate whether a test has passed or failed~\cite{sorokin2026stellar}.
The LLM is instructed to output the score 1 if the response mentions the corresponding warning if necessary (i.e., the test has passed), or if the request is unrelated to the warning, and 0 (i.e., the test has failed) otherwise. We use manually created few-shot examples to guide the generation of reliable assessments.





\subsection{Metrics}

\begin{table*}[t]
\centering
\small
\setlength{\tabcolsep}{4pt}
\renewcommand{\arraystretch}{0.9}

\caption{Results for submitted tools and random as baseline for different systems and manuals, averaged over six runs.}
\label{tab:results-combined-model-multirow}

\begin{tabular}{l l 
                r r r r
                r r r r
                r r r r
                r r r r}
\toprule
\multirow{2}{*}{Model} & \multirow{2}{*}{Tool} 
& \multicolumn{4}{c}{\textbf{\sutone{} -- \manualone{}}}
& \multicolumn{4}{c}{\textbf{\sutone{} -- \manualtwo{}}}
& \multicolumn{4}{c}{\textbf{\suttwo{} -- \manualone{}}}
& \multicolumn{4}{c}{\textbf{\suttwo{} -- \manualtwo{}}} \\
\cmidrule(lr){3-6} \cmidrule(lr){7-10} \cmidrule(lr){11-14} \cmidrule(lr){15-18}
& & \textit{W} & \textit{Rate} & \textit{Cov} & \textit{S}
& \textit{W} & \textit{Rate} & \textit{Cov} & \textit{S}
& \textit{W} & \textit{Rate} & \textit{Cov} & \textit{S}
& \textit{W} & \textit{Rate} & \textit{Cov} & \textit{S} \\
\midrule
\multirow{5}{*}{\gptfouromini} 
&\atlas    & \textbf{0.61} & 0.52 & 0.74 & 0.62 & \textbf{0.64} & 0.51 & 0.92 & \textbf{0.69} & \textbf{0.57} & 0.32 & \textbf{0.89} & \textbf{0.59} & \textbf{0.60} & 0.36 & 0.89 & \textbf{0.62} \\
&\crisp    & 0.42 & 0.32 & 0.50 & 0.42 & 0.46 & 0.32 & 0.68 & 0.49 & 0.34 & 0.15 & 0.61 & 0.37 & 0.35 & 0.17 & 0.69 & 0.41 \\
&\exida    & 0.39 & \textbf{0.65} & 0.60 & 0.55 & 0.41 & \textbf{0.67} & 0.91 & 0.66 & 0.34 & \textbf{0.45} & 0.84 & 0.54 & 0.36 & \textbf{0.45} & 0.85 & 0.55 \\
&\random   & 0.49 & 0.35 & \textbf{0.79} & 0.54 & 0.51 & 0.34 & \textbf{0.94} & 0.60 & 0.38 & 0.23 & 0.88 & 0.50 & 0.40 & 0.24 & \textbf{0.90} & 0.51 \\
&\warnless & 0.51 & 0.58 & 0.78 & \textbf{0.62} & 0.55 & 0.59 & 0.86 & 0.67 & 0.36 & 0.30 & 0.84 & 0.50 & 0.40 & 0.32 & 0.86 & 0.53 \\
\midrule
\multirow{5}{*}{\gptfivechat} 
&\atlas    & \textbf{0.61} & 0.48 & \textbf{0.89} & 0.66 & \textbf{0.64} & 0.59 & 0.86 & 0.69 & \textbf{0.33} & 0.31 & 0.66 & 0.43 & \textbf{0.32} & 0.31 & 0.83 & \textbf{0.49} \\
&\crisp    & 0.45 & 0.27 & 0.71 & 0.47 & 0.48 & 0.31 & 0.65 & 0.48 & 0.26 & 0.35 & \textbf{0.91} & \textbf{0.51} & 0.26 & \textbf{0.38} & 0.60 & 0.41 \\
&\exida    & 0.45 & \textbf{0.73} & 0.87 & \textbf{0.68} & 0.49 & \textbf{0.88} & 0.84 & \textbf{0.73} & 0.21 & \textbf{0.37} & 0.60 & 0.40 & 0.20 & 0.37 & 0.79 & 0.46 \\
&\random   & 0.51 & 0.30 & 0.89 & 0.56 & 0.58 & 0.38 & \textbf{0.91} & 0.63 & 0.20 & 0.20 & 0.67 & 0.36 & 0.21 & 0.21 & \textbf{0.85} & 0.42 \\
&\warnless & 0.54 & 0.53 & 0.87 & 0.65 & 0.59 & 0.68 & 0.86 & 0.71 & 0.18 & 0.22 & 0.58 & 0.33 & 0.20 & 0.25 & 0.81 & 0.42 \\
\bottomrule
\end{tabular}
\end{table*}

We used the following metrics for the evaluation of the tools: 


\textit{Number of warnings ignored ($W$)}. To assess the robustness of the LLM-based application, we want to assess how many different warnings in the manual have been ignored.

\textit{Failure Rate ($Rate$)}. The failure rate is defined as the ratio of failed test inputs to the total number of generated test inputs. This metric accounts for differences among test generation approaches in the time required to produce individual test inputs~\cite{sorokin2026stellar}. In addition, it accommodates variability in execution conditions across tools, e.g., due to nondeterministic behavior introduced by LLM-based calls.

\textit{Failure Coverage ($Cov$)}: This metric serves as an indicator of how well a test generator can cover existing failures~\cite{surrogate2024biagiola}. In particular, we focus here on the input space coverage, e.g., on the coverage of failure revealing test inputs.
To approximate the coverage value, we first aggregate all failing test inputs from all test approaches to \textit{assess} the total failure space. Then, we apply k-Means clustering with the silhouette method~\cite{surrogate2024biagiola} to obtain existing failure clusters. We then merge all failing test cases produced by the respective test generator and assign each failure to one of the clusters computed in the previous step. Based on this assignment, we determine the number of clusters covered by at least one failing test case and compute the corresponding coverage score~\cite{sorokin2026stellar}. We configure the min number of clusters for the silhouette method by using default values~\cite{surrogate2024biagiola} and repeat the clustering and cluster assignment 10 times.

\textit{Overall Score ($S$).} We combine all metric results into one score to rank the testing approaches. For the number of warnings ignored, we normalize the score based on the total number of warnings in the manual $W'$. Finally, we use a weighted sum with equal weights to combine the scores as follows: 
\begin{align*}
    S = \frac{1}{3} \left( W' + \mathrm{Rate} + \mathrm{Cov} \right).
\end{align*} 
We exclude the number of failures from the overall ranking, as an objective normalization of this metric is not feasible. Moreover, the failure rate metric already implicitly accounts for the number of failures. The corresponding weights are selected equally, as no prioritization between metrics is applied.

\subsection{Tools}
In total, four tools were submitted to the competition. We briefly describe the main characteristics of each tool. More information can be
found in the corresponding reports:

\textbf{ATLAS}~\cite{ATLAS-deeptest26} combines priority-based warning selection with human-like perturbations to identify diverse failures. The first component assigns weights to already explored warnings to prioritize the exploration of new failures, while the second component applies word-level perturbations such as adding filler words  (e.g., ``hm'' or ``uhm'') or deletion of words ~\cite{sorokin2026stellar} to generate diverse test inputs.

The \textbf{Exida} test generator~\cite{Exida-deeptest26} is an automated tool that creates test inputs based on three steps: first, it generates a realistic situation that requires input from the manual, second, it formulates a user intent, and third, it crafts a natural question with the support of an LLM to trigger specific warnings. It uses Jaccard similarity as the diversity mechanism to avoid repetitive questions and ensure variety between generated tests.

\textbf{Warnless}~\cite{Warnless-deeptest26} employs an LLM-based utterance generator and a probability-based warning selection strategy in which the selection probability is dynamically adjusted according to the proportion of successful trials inducing system failures. Warnings that more frequently lead to failures are assigned higher probabilities and are thus more likely to be selected again.

\textbf{CRISP}~\cite{CRISP-deeptest26} uses predefined phrases to create natural, constraint-compliant, and realistic in-car utterances. \crisp employs a so-called Risk Context Navigator, which selects relevant safety-related conditions (e.g., adverse weather), and a Generator, which combines the chosen context with concise language to produce requests that lead to brief LLM responses.

\textbf{Random}~\cite{repo} is a baseline test generation approach that employs an LLM instructed to produce human-like utterances based on a randomly sampled set of warnings included in the manual.%

\subsection{Test Subjects}

For the evaluation of the testing tools, we used two different LLM-based assistants \textit{\sutone}, an open-source implementation provided to the participants during the competition, and \textit{\suttwo}, an industrial system provided by BMW. Two different manuals \textit{\manualone} and \textit{\manualtwo} from different vehicle models were used in the backend of the SUT to evaluate the generalization capabilities of the tools.

We evaluated all test subjects with \gptfouromini as well as with \gptfivechat in the backend for the processing of the request with a max token size of 1500 and temperature 0.

\subsection{Procedure}

Before the tool evaluation, we applied the following procedure: first, we benchmarked different cloud-based LLMs to be used as oracles for the evaluation of executed tests. Our benchmark included 1000 synthetically (with \gptfouro generated) and human-based validated request and answer pairs.
Among the tested LLMs, \gptfouromini yielded the best balance between accuracy and precision (F1-score = 0.824), execution time (0.88s/request), and costs (0.2\$/1000 requests). Based on these results, we selected this LLM to be used in the pipeline for tool evaluation.
For the final evaluation for each tool, we selected a search budget of 2 hours as suggested by BMW experts. We recorded all generated test inputs, test outputs as well as judge results for the subsequent analysis. For tools that required a specific configuration, such as LLM usage, we selected the default setup provided by the tool.

%% file: content/results.tex
\section{Results}

The results for each metric and tool per evaluated SUT averaged over six runs\footnote{
Experiments were run on Ubuntu featuring a sixteen-core Intel Xeon Gold 5222 3.80GHz CPU, with 64 GB of RAM and an NVIDIA RTX A4000 GPU.}  and different LLMs are provided in \autoref{tab:results-combined-model-multirow}.

\begin{figure}[t]
    \centering
    \begin{subfigure}[b]{0.485\linewidth}
        \centering
        \includegraphics[width=\linewidth, trim=0 65 0 5, clip]{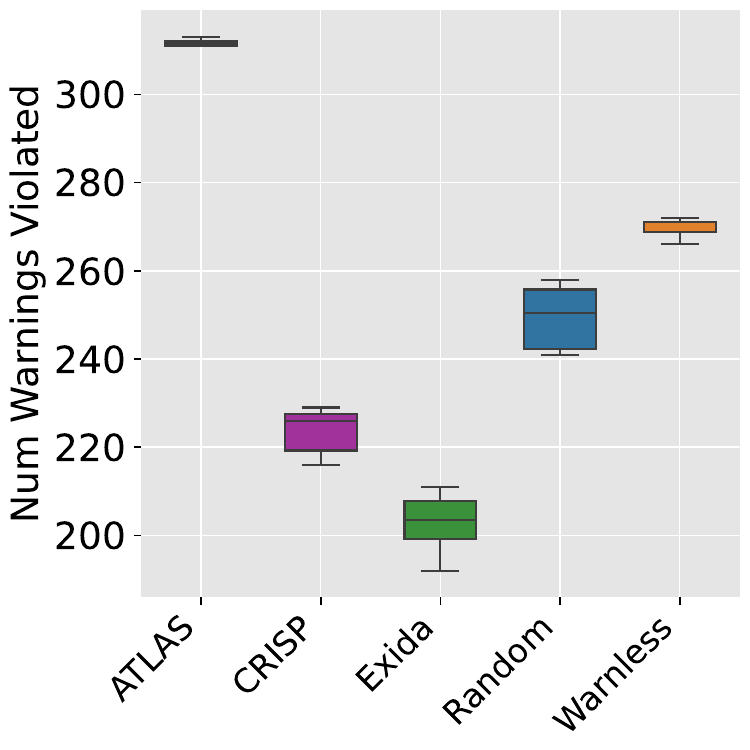}        \label{fig:sub3}
    \end{subfigure}%
    \hspace{0.02\linewidth}
    \begin{subfigure}[b]{0.485\linewidth}
        \centering
     \includegraphics[width=\linewidth, trim=0 65 0 5, clip]{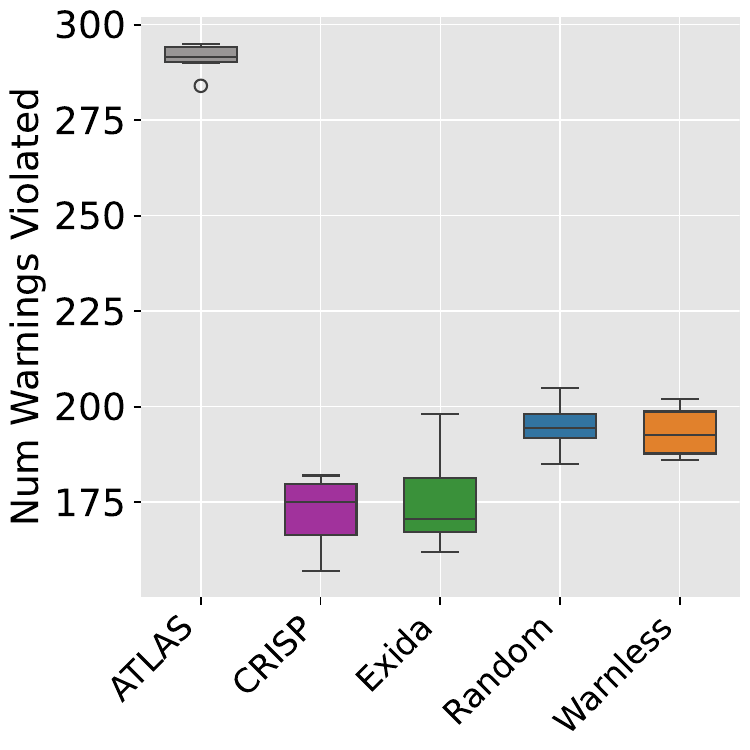}          \label{fig:sub4}
    \end{subfigure}
  
    \begin{subfigure}[b]{0.485\linewidth}
        \centering
     \includegraphics[width=\linewidth, trim=0 65 0 10, clip]{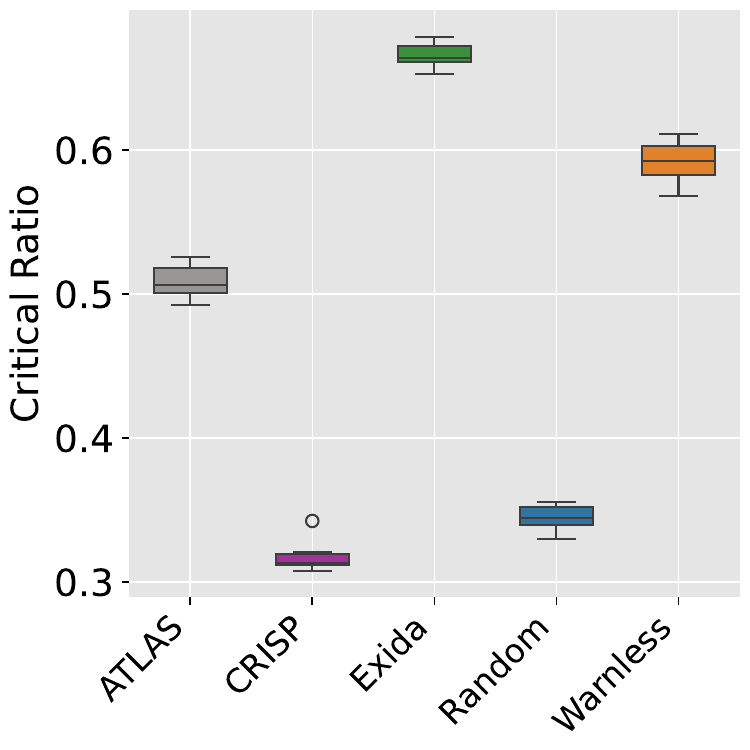}
     \label{fig:sub1}
    \end{subfigure}%
    \hspace{0.02\linewidth} 
    \begin{subfigure}[b]{0.485\linewidth}
        \centering
     \includegraphics[width=\linewidth, trim=0 65 0 5, clip]{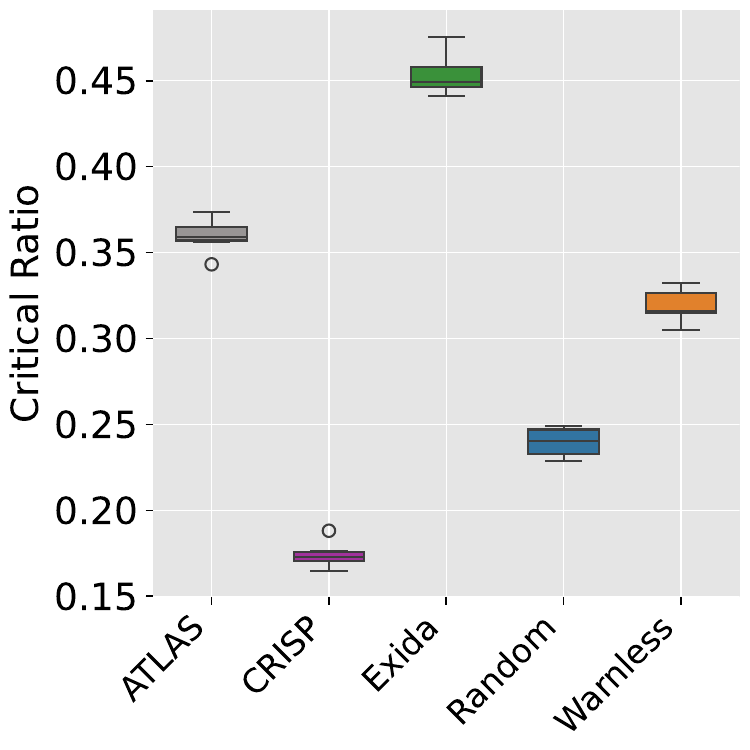}          \label{fig:sub2}
    \end{subfigure}
    \vspace{0.3cm}  
    \begin{subfigure}[b]{0.485\linewidth}
        \centering
        \includegraphics[width=\linewidth, trim=0 0 0 10, clip]{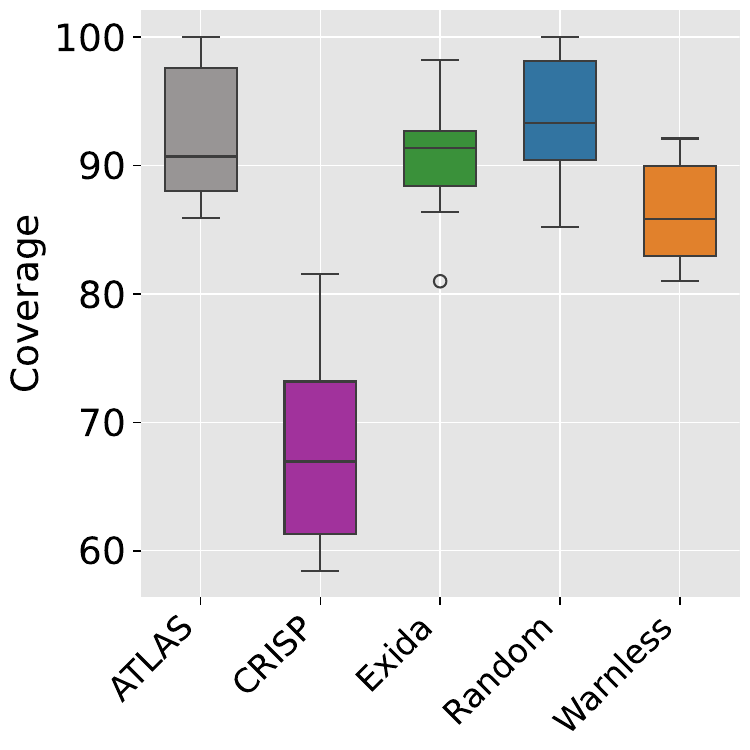}        \label{fig:sub3}
    \end{subfigure}%
    \hspace{0.02\linewidth}
    \begin{subfigure}[b]{0.485\linewidth}
        \centering
     \includegraphics[width=\linewidth, trim=0 0 0 4, clip]{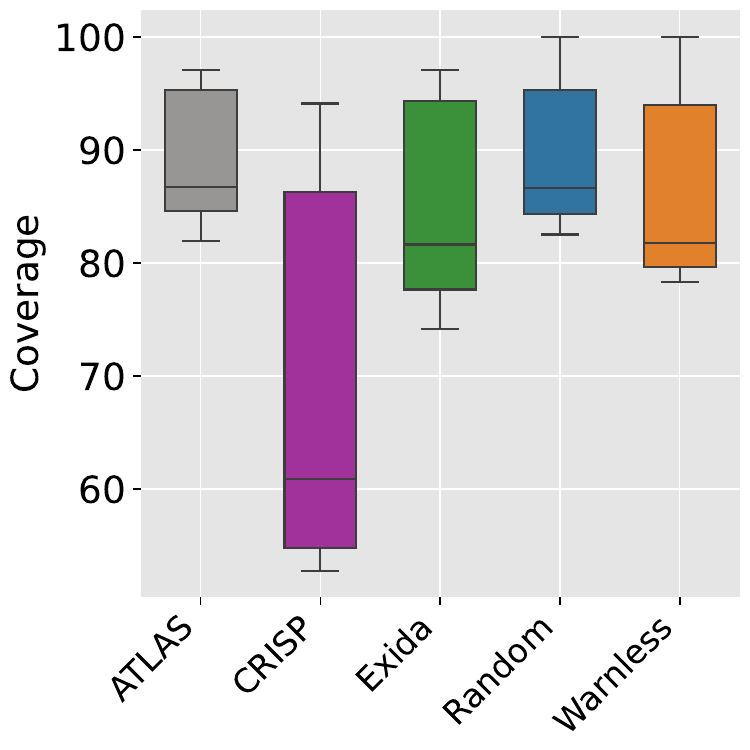}          \label{fig:sub4}
    \end{subfigure}
    \caption{Metric results for \sutone (left) and \suttwo (right) with manual \manualtwo and LLM \gptfouromini  over 6 runs.}
    \label{fig:boxplots-results}
\end{figure}

Concerning the failure rate, \exida reached the highest score of 57\%, followed by \warnless with 43\%. While \crisp achieved the lowest rate, it could identify 898 failures, being the highest number of failures, followed by \warnless with a count of 778. The highest warning and failure coverage were achieved by \atlas, with scores of 0.54 and 0.83, respectively. \warnless followed with a failure coverage of 0.78, while \crisp achieved a ratio of failed warnings of 0.38.
In the overall ranking, \atlas achieved the highest score of 0.59, followed by \exida with 0.57, \warnless with 0.55, and \crisp with 0.44. The randomized baseline approach reached a score of 0.51. 
\autoref{fig:boxplots-results} shows boxplot results for \sutone and \suttwo with \manualtwo and \gptfouromini. The lowest variation across metrics and tools is observed for the failure rate, while the highest variation occurs for the coverage scores. For some tools, coverage variation is higher for \suttwo than for \sutone. For the remaining metrics, no significant differences are observable.

Regarding the SUT and LLM variation, the results obtained with \manualone were similar to those observed for \manualtwo. Across nearly all configurations, testing the industrial assistant \suttwo yield lower overall scores than the open-source system \sutone. For \suttwo, the use of \gptfivechat led to a measurable improvement compared to \gptfouromini, whereas no significant effect of the model choice was observed for \sutone.

\renewcommand{\arraystretch}{0.95} 

Examples of generated utterances related to a warning about the correct installation of a child seat are shown in \autoref{tab:examples-results}. \atlas explicitly highlights the urgency of the request and includes a filler word, whereas \exida frames the utterance within a concrete scenario. \warnless and \random specifically target the backrest, while \crisp produces a broader question that includes environmental context. Extended results regarding the variation of the different metric scores for each SUT and manual configuration as well as generated utterances are available online~\cite{repo}.

%% file: content/conclusion.tex
\section{Conclusions}

The DeepTest 2026 Testing Competition focused on benchmarking an LLM-based Assistant for retrieving information from an owner's manual. In total, four tools and 10 participants were competing in testing one industrial and one open-source implementation an automotive assistant with two different manual configurations. Overall, \atlas achieved the highest average score when leveraging human-like perturbations, while \exida performed best in optimizing the failure rate, and \crisp identified the highest number of failures. The results demonstrate that different testing approaches can effectively benchmark LLM-based applications, highlighting the importance of employing diverse methodologies to evaluate LLM-based assistants.

%% file: paper.bib
@misc{repo,
	title        = {Replication Package},
	author       = {Sorokin, Lev and Vasilev, Ivan and Pasini, Samuele},
	howpublished = {\url{https://github.com/deeptest-competition}}
}

@inproceedings{2025-Giebisch-IV,
	title        = {Automated Factual Benchmarking for In-Car Conversational Systems using Large Language Models},
	author       = {Giebisch, Rafael and Friedl, Ken E. and Sorokin, Lev and Stocco, Andrea},
	year         = 2025,
	booktitle    = {Proceedings of the 36th IEEE Intelligent Vehicles Symposium},
	series       = {IV '25}
}

@article{gu2024surveyllmasajudge,
	title        = {A Survey on LLM-as-a-Judge},
	author       = {Jiawei Gu and Xuhui Jiang and Zhichao Shi and Hexiang Tan and Xuehao Zhai and Chengjin Xu and Wei Li and Yinghan Shen and Shengjie Ma and Honghao Liu and Yuanzhuo Wang and Jian Guo},
	year         = 2024,
	journal      = {arXiv preprint arXiv: 2411.15594}
}

@article{surrogate2024biagiola,
	title        = {Testing of Deep Reinforcement Learning Agents with Surrogate Models},
	author       = {Biagiola, Matteo and Tonella, Paolo},
	year         = 2024,
	journal      = {ACM Trans. Softw. Eng. Methodol.},
	volume       = 33,
	number       = 3,
	issn         = {1049-331X},
	issue_date   = {March 2024},
	articleno    = 73,
	numpages     = 33
}

@inproceedings{ahmed2025rag,
	title        = {{Quality Assurance for LLM-RAG Systems: Empirical Insights from Tourism Application Testing}},
	author       = {Ahmed, Bestoun S. and Baader, Ludwig Otto and Bayram, Firas and Jagstedt, Siri and Magnusson, Peter},
	year         = 2025,
	booktitle    = {ICSTW},
	volume       = {},
	issn         = {2159-4848}
}

@misc{habicht2025benchmarkingcontextualunderstandingincar,
      title={Benchmarking Contextual Understanding for In-Car Conversational Systems}, 
      author={Philipp Habicht and Lev Sorokin and Abdullah Saydemir and Ken E. Friedl and Andrea Stocco},
      year={2025},
      eprint={2512.12042},
      archivePrefix={arXiv},
      primaryClass={cs.CL}, 
}

@inproceedings{sorokin2026stellar,
  title     = {STELLAR: A Search-Based Testing Framework for Large Language Model Applications},
  author    = {Sorokin, Lev and Vasilev, Ivan and Friedl, Ken E. and Stocco, Andrea},
  booktitle = {Proceedings of the 33rd IEEE International Conference on Software Analysis, Evolution and Reengineering},
  year      = {2026},
  publisher = {IEEE},
}

@inproceedings{ATLAS-deeptest26,
  title     = {ATLAS: Adaptive Test Learning And Selection},
  author    = {Antonio Pedro Santos Alves and Marcos Kalinowski},
  booktitle = {Proceedings of the Seventh International Workshop on Deep Learning for Testing and Testing for Deep Learning ({DeepTest} 2026), co-located with the {IEEE/ACM} International Conference on Software Engineering},
  year      = {2026}
}

@inproceedings{Exida-deeptest26,
  title     = {Multistage Prompt Decomposition for
Failure-Inducing Tests},
  author    = {Kaan-Gueney Keklikci and Gaia Peressini and Dimitrij Krepis},
  booktitle = {Proceedings of the Seventh International Workshop on Deep Learning for Testing and Testing for Deep Learning ({DeepTest} 2026), co-located with the {IEEE/ACM} International Conference on Software Engineering ({ICSE} 2026)},
  year      = {2026}
}

@inproceedings{Warnless-deeptest26,
  title     = {{Warnless at the DeepTest 2026 Tool Competition}},
  author    = {Song Qunying and Yuan Gao and Roberto Brusnicki and Federica Sarro},
  booktitle = {Proceedings of the Seventh International Workshop on Deep Learning for Testing and Testing for Deep Learning ({DeepTest} 2026), co-located with the {IEEE/ACM} International Conference on Software Engineering ({ICSE} 2026)},
  year      = {2026}
}

@inproceedings{CRISP-deeptest26,
  title     = {CRISP: Contextual Risk-Driven Input Structuring for Probing},
  author    = {Selhan Berber},
  booktitle = {Proceedings of the Seventh International Workshop on Deep Learning for Testing and Testing for Deep Learning ({DeepTest} 2026), co-located with the {IEEE/ACM} International Conference on Software Engineering ({ICSE} 2026)},
  year      = {2026}
}

@misc{enchant,
  title        = {Enchant: A Generic Spell Checking Library},
  author       = {Reuben Thomas},
  howpublished = {\url{https://rrthomas.github.io/enchant/}},
  note         = {Accessed: 2026-01-25},
  year         = {2026}
}

@misc{bmw2026_alexa,
  author       = {{BMW Group Press Club}},
  title        = {A Milestone for Human–Vehicle Interaction: BMW Intelligent Personal Assistant Expanded to Include Amazon Alexa Technology},
  note         = {Accessed: 2026-01-25},
  year         = {2026}
}
